\documentclass[sigconf]{acmart}

\newcommand{\CudaSimGpuOverCpuSpeedup}{1024}
\newcommand{\CudaSimDataCollectionMultiCoreCpuSpeedup}{4}

\newcommand{\CudaSimSimulatorComplexityMaxSpeedup}{1024}

\newcommand{\CudaSimBestCaseFusionSpeedupMax}{1.729057454209506}

\newcommand{\CudaSimJaxDataCollectionSpeedupJaxOverTorch}{13.391258389034308}
\newcommand{\CudaSimJaxDataCollectionSpeedupSteps}{53.757822636636035}

\newcommand{\CudaSimSimulationJaxAmortizedCpuOverheadSpeedup}{83.34774829937089}
\newcommand{\CudaSimSimulationKernelFusionAndNoCpuOverheadSpeedup}{112.76804635714576}
\newcommand{\CudaSimSimulationKernelFusionSpeedup}{11.305685226074862}

\input{tex/custom_imports}

\newcommand{\kernelFusion}{kernel fusion\xspace}
\newcommand{\KernelFusion}{Kernel fusion\xspace}

\newcommand{\simKernelFusion}{simulator kernel fusion\xspace}
\newcommand{\SimKernelFusion}{Simulator kernel fusion\xspace}
\newcommand{\SimKernelFusionCamel}{Simulator Kernel Fusion\xspace}

\newcommand{\CudaSimXLA}{XLA\xspace}

\newcommand{\gpuVec}{GPU vectorization\xspace}
\newcommand{\GpuVec}{GPU Vectorization\xspace}
\newcommand{\GpuVecCamel}{GPU Vectorization\xspace}

\newcommand{\PaperTitle}{Optimizing Data Collection in Deep Reinforcement Learning}

\newcommand{\CiteExtra}[1]{\xspace}

\NewDocumentCommand{\codeword}{v}{\texttt{#1}}

\newcommand{\irect}[1]{%
  \tikz[baseline=(char.base)]\node[anchor=south west, draw,rectangle, rounded corners=0.5mm, inner sep=1.5pt, minimum size=3mm,outer xsep=0pt,column sep=0pt,row sep=0pt,
    text height=2.2mm](char){#1} ;}

\newcounter{FndQualCounter}

\newcounter{FndSurpCounter}

\newcounter{FndCounter}
\newcommand{\Fnd}[1]{%
  \refstepcounter{FndCounter}
  \label{#1}
  \irect{\textit{F}.\theFndCounter}\xspace
}
\newcommand{\RefFnd}[1]{%
  \irect{\hyperref[#1]{\textit{F}.\ref{#1}}}\xspace
}

\newcommand{\IntroRefFnd}[1]{}

\newcommand{\EvalRefFnd}[1]{}

\newenvironment{finding}[1]
{
    \begin{tcolorbox}[size=fbox,after skip=0.4\baselineskip]
    \Fnd{#1} 
    \bfseries
}
{
    \end{tcolorbox}
}

\newcommand{\skeleton}[2]{
  \ifx\SkeletonOutline\undefined
    #2
  \else
    #1
  \fi %
}
\newcommand{\BulletPoint}[1]{
  $\bullet$ #1 \\
}

\newtcbox{\legendbox}[1]{on line, colback=#1!10!white, colframe=#1!50!black, size=fbox, fontupper=\sffamily\scriptsize, arc=0mm, boxsep=2pt}

\newtcbox{\barbox}[1]{on line, colback=#1!30!white, colframe=#1!50!black, size=fbox, fontupper=\sffamily\scriptsize, arc=0mm, boxsep=2pt}

\AtBeginDocument{%
  }

\setcopyright{acmcopyright}
\copyrightyear{2022}
\acmYear{2022}
\acmDOI{}

\acmConference[MLBench 2022]{Third Workshop on Benchmarking Machine Learning Workloads on Emerging Hardware}
  {September 1, 2022}
  {Santa Clara, California, USA}
\acmPrice{}
\acmISBN{}

\begin{document}

\newcommand{\AsTimes}[1]{\pgfmathparse{#1}\num[round-mode=places,round-precision=1]{\pgfmathresult}$\times$}
\newcommand{\AsPercent}[1]{\pgfmathparse{100*#1}\num[round-mode=places,round-precision=1]{\pgfmathresult}\%}

\title{\PaperTitle}

\author{James Gleeson}
\email{jgleeson@cs.toronto.edu}
\affiliation{%
  \institution{University of Toronto, Vector Institute}
  \city{Toronto}
  \state{Ontario}
  \country{Canada}
}

\author{Daniel Snider}
\email{dans@cs.toronto.edu}
\affiliation{%
  \institution{University of Toronto, Vector Institute}
  \city{Toronto}
  \state{Ontario}
  \country{Canada}
}

\author{Yvonne Yang}
\email{yvon.yang@mail.utoronto.ca}
\affiliation{%
  \institution{University of Toronto}
  \city{Toronto}
  \state{Ontario}
  \country{Canada}
}

\author{Moshe Gabel}
\email{mgabel@cs.toronto.edu}
\affiliation{%
  \institution{University of Toronto}
  \city{Toronto}
  \state{Ontario}
  \country{Canada}
}

\author{Eyal de Lara}
\email{delara@cs.toronto.edu}
\affiliation{%
  \institution{University of Toronto}
  \city{Toronto}
  \state{Ontario}
  \country{Canada}
}

\author{Gennady Pekhimenko}
\email{pekhimenko@cs.toronto.edu}
\affiliation{%
  \institution{University of Toronto, Vector Institute}
  \city{Toronto}
  \state{Ontario}
  \country{Canada}
}

\renewcommand{\shortauthors}{Gleeson et al.}

\begin{abstract}

Reinforcement learning (RL) workloads take a notoriously long time to train due to the large number of samples collected at runtime from simulators.
Unfortunately, cluster scale-up approaches remain expensive, and commonly used CPU implementations of simulators induce high overhead when switching back and forth between GPU computations.
We explore two optimizations that increase RL data collection efficiency by increasing GPU utilization: (1) \emph{\gpuVec:} parallelizing simulation on the GPU for increased hardware parallelism, and (2) \emph{\simKernelFusion:} fusing multiple simulation steps to run in a single GPU kernel launch to reduce global memory bandwidth requirements. 
We find that \gpuVec can achieve up to
$\CudaSimGpuOverCpuSpeedup \times$
speedup over commonly used CPU simulators. 
We profile the performance of different implementations and show that for a simple simulator,  
ML compiler implementations (\CudaSimXLA) of \gpuVec outperform a DNN framework (PyTorch) by \AsTimes{\CudaSimJaxDataCollectionSpeedupJaxOverTorch} 
by reducing CPU overhead from repeated Python to DL backend API calls.
We show that \simKernelFusion speedups with a simple simulator are \AsTimes{\CudaSimSimulationKernelFusionSpeedup} and increase by up to 
$\CudaSimSimulatorComplexityMaxSpeedup \times$
as simulator complexity increases in terms of memory bandwidth requirements.
We show that the speedups from \simKernelFusion are orthogonal and combinable with \gpuVec, leading to a multiplicative speedup.  

\ifx\AddPageCount\undefined
\else
{\hfill\color{red} $\Rightarrow$ \textbf{Num pages = \pageref*{page:last-page}} $\Leftarrow$}
\fi

\end{abstract}



\maketitle

\section{Introduction}
\label{sec:intro}

\skeleton{
    \BulletPoint{RL training takes a long time because it requires collecting many samples from simulators.}
    \BulletPoint{Data collection is embarassingly parallel and can be sped up using a cluster (e.g., AlphaZero~\cite{silver2018general}), but it is too costly for most many people.}
}{
Reinforcement learning (RL) workloads take a notoriously long time to train due to the large number of samples collected at runtime from simulators.
Recent works address this problem 
by running multiple simulators in parallel across a cluster of accelerator-equipped machines.
For example, 
AlphaZero~\cite{silver2018general} used 5000 TPUs to perform self-play in parallel reducing training time to 13 days.
However, at an hourly cloud pricing of $\$6.50/\text{hour}$ for an on-demand v1 TPU~\cite{tpu-v1-pricing-2018} used at the time, that brings the cost of each trained model to $\$10,140,000$.
Hence, scaling up RL training is economically impractical outside of large-scale industrial research projects.
}

\skeleton{
    \BulletPoint{RL-Scope showed that RL training workloads are inefficiently implemented, with most of their time spent CPU-bound (e.g., CUDA API calls), not GPU-bound.}
    \BulletPoint{Simulation time from CPU-based simulators non-negligible at $38.1\%$.}
    \BulletPoint{RL frameworks need to find ways to better utilize GPUs to increase throughput.}
}{

A recent survey of RL training workloads~\cite{gleeson2021rlscope} demonstrated that the low GPU utilization of RL workloads is caused by data collection, which manifests in two ways.
First, CPU simulation time takes up a large amount of training time, with at least $38.1\%$ of training time spent in simulation across popular robotics simulators.
Second, a large amount of CPU time originates from overheads induced by switching between CPU-based simulation and GPU-based inference.
CUDA API calls alone account for $3.6\times$ as much time on average as the GPU kernel execution in both PyTorch \cite{pytorch} and TensorFlow \cite{abadi2016tensorflow} implementations of an RL algorithm.
Hence, to optimize data collection in RL frameworks, GPU utilization must be increased.

}

\skeleton{
    \BulletPoint{This paper explores two potential optimizations for increasing GPU utilization: (1) \gpuVec, and (2) \kernelFusion.}
}{
In this paper, we explore two potential optimizations for increasing GPU utilization during the time-consuming data collection phase of RL training workloads: (1) \emph{\gpuVec}, and (2) \emph{\simKernelFusion}.
This work is preliminary, as it focuses on a simple simulator that eased implementation efforts across different ML frameworks thereby enabling a more in-depth analysis.
Additional simulators will be explored in future work (\Cref{sec:future-work}).

}

\skeleton{
    \BulletPoint{\emph{\GpuVec} co-locates simulation and inference on the GPU by expressing simulation computations in DL frameworks.}
    \BulletPoint{This allows us to run more simulators concurrently using GPU parallelism, and reduce CPU overheads (e.g., CUDA API calls, CPU$\leftrightarrow$GPU data copies).}
    \BulletPoint{We speed up data collection a lot: up to \AsTimes{\CudaSimGpuOverCpuSpeedup} with PyTorch and \CudaSimXLA.}
}{

\emph{\gpuVec} exploits the massive hardware parallelism of GPUs to run $N$ simulator instances in parallel.
This leads to performance benefits from a greater amount of simulators that can run in parallel and from reduced CPU overheads (e.g., data copying, API calls) when switching to and from the GPU.
Using both DL framework (PyTorch) and ML compiler (\CudaSimXLA \cite{xla}) approaches, we are able to achieve up to \AsTimes{\CudaSimGpuOverCpuSpeedup} speedup in data collection. 
Prior work \cite{brax2021github} also observed that a GPU implementation can have large speedups, but limited their comparisons to only a single ML framework, and only compared against an inefficient single-core CPU implementation.
In contrast, we compare and contrast the full gamut of commonly used and high-performance implementations of vectorization.
We show that using multiple cores with C++ only accounts for a 
\AsTimes{\CudaSimDataCollectionMultiCoreCpuSpeedup} 
speedup over the single-core OpenAI implementation.
We compare different GPU implementations and show that \CudaSimXLA can achieve \AsTimes{\CudaSimJaxDataCollectionSpeedupJaxOverTorch} speedup over PyTorch.
Profiling reveals that XLA amortizes CPU overheads by launching all data collection GPU kernels in a constant number of Python$\rightarrow$XLA API calls regardless of the number of steps, whereas PyTorch requires a greater number of 
Python$\rightarrow$PyTorch 
API calls as data collection steps increase.

}

\skeleton{
    \BulletPoint{\emph{Inference/simulation kernel fusion} fuses the simulation GPU kernel with the first neural network layer of the DNN model.}
    \BulletPoint{Simulation is ideally suited to kernel fusion, since kernels are short and operations are simple row-wise computations.}
    \BulletPoint{If data is kept entirely in registers, simulation alone can be sped up by \AsTimes{\CudaSimSimulationKernelFusionAndNoCpuOverheadSpeedup}.}
    \BulletPoint{Unfortunately, simulation GPU time is dominated by inference GPU time, so potential speedups from fusion are limited to \AsTimes{\CudaSimBestCaseFusionSpeedupMax} overall.} 
    \BulletPoint{We recommend developers adopt \gpuVec by expressing simulation in DL frameworks.} 
}{
\emph{\KernelFusion} is a GPU optimization that benefits from (1) reduced kernel launch overhead from fewer kernel launches, and (2) increased cache efficiency by avoiding device memory transfers. 
Many common robotic physics simulators can be modeled as rigid body simulations that are parallelized over the joint states of the simulation \cite{brax2021github}.
Since these simulation GPU kernels are short, this makes them an attractive target for kernel fusion.
To demonstrate this, we fused multiple simulation steps in the simple cartpole \cite{cartpole} simulator and achieved \AsTimes{\CudaSimSimulationKernelFusionSpeedup} speedup.
For increasingly complex simulators, we show that speedups of kernel fusion are larger for more memory bandwidth bound simulators since fusion reduces global memory transfers.
In particular, kernel fusion can have $8 - 1024\times$ speedup depending on how memory bandwidth bound the simulator is.
The speedup from kernel fusion is independent of the number of parallel simulators used, so kernel fusion can be combined with \gpuVec to achieve massive simulation speedups.
}

In summary, our contributions are:
\begin{itemize}[nosep, leftmargin=*]

    \item 
    We thoroughly compare the performance limitations of \gpuVec implementations.
    Both ML compiler and deep neural network (DNN) frameworks are up to $\CudaSimGpuOverCpuSpeedup\times$ faster as parallel environments saturate.
    However, ML compiler approaches out-perform DNN framework approaches by \AsTimes{\CudaSimJaxDataCollectionSpeedupJaxOverTorch} for smaller parallel environments configurations by amortizing CPU overheads.

    \item 
    We show that \simKernelFusion achieves \AsTimes{\CudaSimSimulationKernelFusionSpeedup} speedup for a real simulator and that the maximum speedup increases up to $1024\times$ for memory bandwidth bound simulators of increasing complexity.  

    \item 
    We demonstrate that \simKernelFusion speedups and \gpuVec speedups are independent, and both can be combined for massive multiplicative benefits.

\end{itemize}

\section{Background}
\label{sec:background}

\newcommand{\FigRlTrainingLoopTxtColor}{Plum}
\newsavebox{\FigRlTrainingLoopLargestFigure}
\begin{figure*}[t]
     \centering
     \savebox{\FigRlTrainingLoopLargestFigure}{\includegraphics[width=0.68\textwidth]{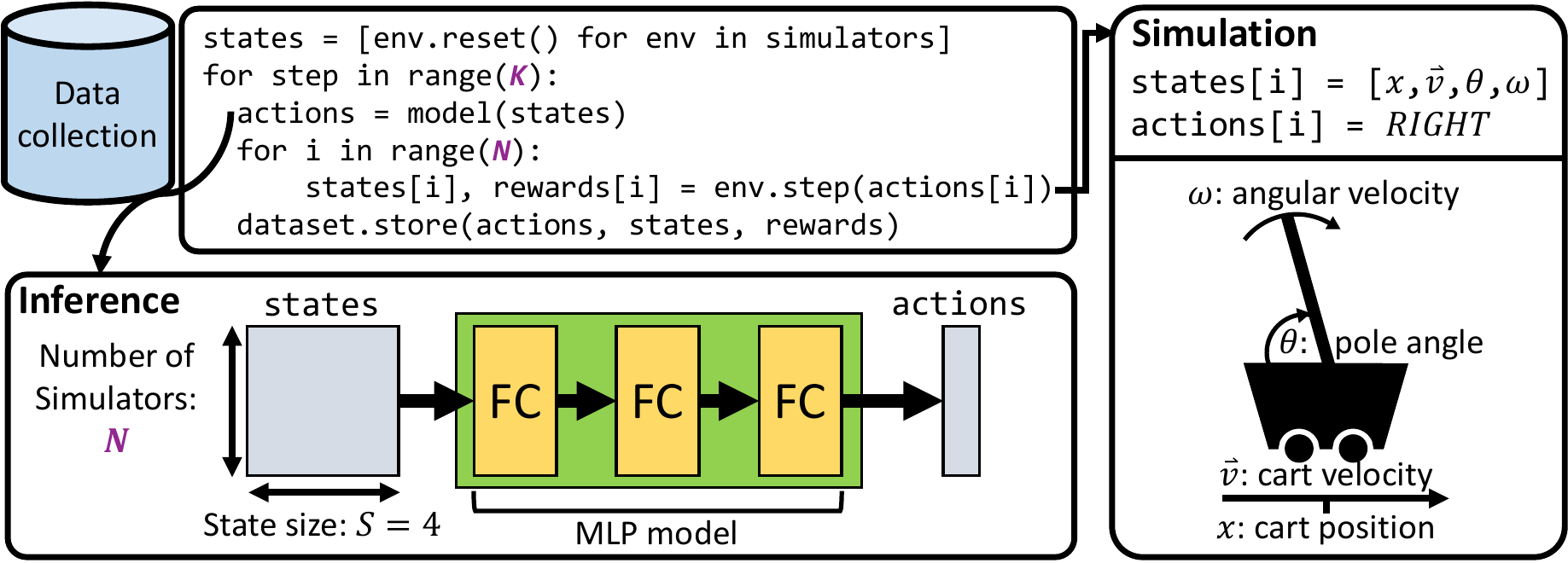}}
     \begin{subfigure}[t]{0.28\textwidth}
         \centering
         \raisebox{\dimexpr.5\ht\FigRlTrainingLoopLargestFigure-.5\height}{
            \includegraphics[width=\textwidth]{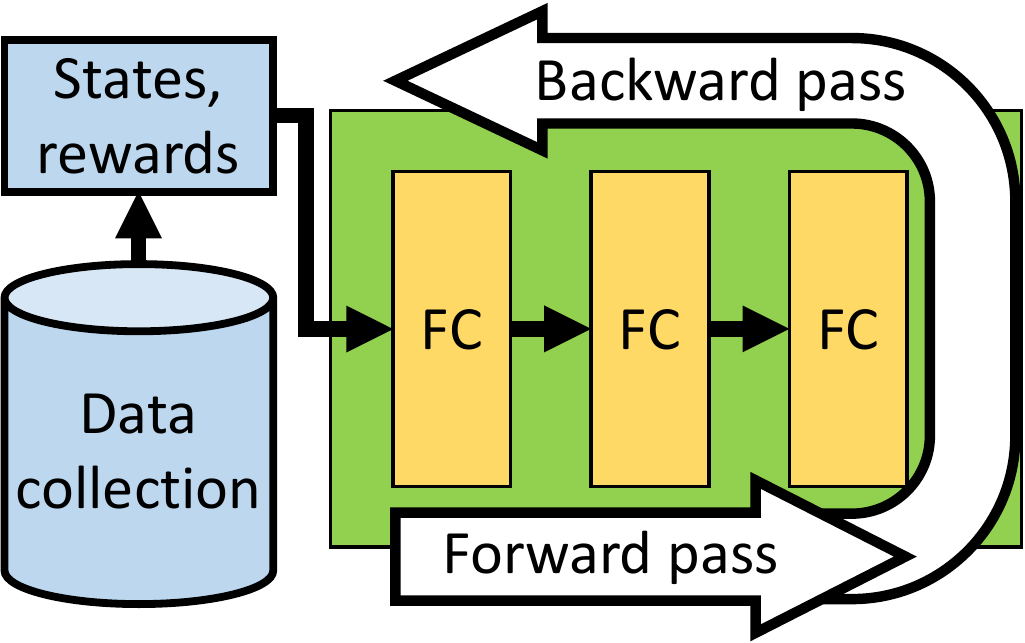}
         }
         \caption{\textbf{RL training:} RL training uses the backpropagation algorithm similar to supervised learning, except that the training data must be collected at runtime.}
         \label{fig:rl-training}
     \end{subfigure}
     \hfill
     \begin{subfigure}[t]{0.68\textwidth}
         \centering
         \usebox{\FigRlTrainingLoopLargestFigure}
         \caption{\textbf{Data collection:} data collection is a simulation/inference loop.  Common RL training frameworks use CPU-based simulator implementations (e.g., OpenAI gym), and assemble a single minibatch to perform inference on the GPU.  Data collection hyperparameters (\textcolor{\FigRlTrainingLoopTxtColor}{\textbf{\emph{N}}} simulator instances, \textcolor{\FigRlTrainingLoopTxtColor}{\textbf{\emph{K}}} simulator steps) are shown.}
         \label{fig:data-collection}
     \end{subfigure}
     \caption{\textbf{RL training loop:} A high-level breakdown of how RL training is performed.}
     \label{fig:rl-training-and-data-collection}
\end{figure*}

\skeleton{
    \BulletPoint{We summarize RL training and data collection using the cartpole simulator, which is a best-case scenario for the two optimizations we explore.}
}{
We provide a high-level overview of the RL training procedure with a focus on data collection, and ML frameworks used to implement it.
We describe a simple simulator that we use to explore
\gpuVec and \simKernelFusion optimizations and demonstrate that it is representative of robotics physics simulators commonly used in RL.
}

\subsection{RL Training}
\label{sec:rl-training}

\skeleton{
    \BulletPoint{RL uses backpropagation to learn (\Cref{fig:rl-training}), but must collection training data at runtime through simulation in a process called data collection.}
    \BulletPoint{Data collection iteratively performs simulation and inference for a certain number of \emph{steps} to ensure a policy is learned, and with \emph{N} simulator instances to speed up collection (\Cref{fig:data-collection}).}
    \BulletPoint{Most data collection implementations only use CPUs, and are inefficiently implemented in Python.}
}{
The RL training procedure uses backpropagation (shown in \Cref{fig:rl-training}), 
and collects the training dataset at run-time by interacting with a simulator.
The data collection process (\Cref{fig:data-collection}) consists of a simulation/inference loop, whereby the state of the simulator is fed into the model learned so far to determine which action to take.
The selected action determines the next state of the simulator and the resulting reward, and the reward is used to form labels for the collected data.

The data collection loop runs for a pre-determined number of steps \emph{K} (i.e., a hyperparameter).
Different ML frameworks implement the data collection loop differently, which we will show has performance implications.
DNN frameworks (PyTorch) execute $K$ separate DL backend API calls for each step, 
whereas ML compilers (XLA) can condense the entire loop into a single DL backend API call.
Once enough data is collected from the simulator, the model is updated using the backpropagation algorithm, which completes a training epoch for the RL algorithm.

To accelerate data collection, RL training frameworks will run multiple simulator instances \emph{N}.
Simulators are typically CPU-based (e.g., Mujoco \cite{todorov2012mujoco}, PyBullet \cite{pybullet}, OpenAI gym \cite{brockman2016openai}). 
Some RL training frameworks use multiple CPU cores to parallelize simulators \cite{ray}, whereas others opt for flexibility and use only a single CPU core\footnote{Python RL frameworks \cite{stable-baselines} opt to use only a single CPU core due to inefficiencies in Python's shared memory multi-threading implementation \cite{rl-zoo-single-core}}; in our evaluation we consider both.
Multiple simulator states are combined into a single minibatch that is fed to inference running on the GPU.
Training continues until a pre-determined number of training epochs has completed, after which the average reward per episode will converge to a maximum value.
}

\subsection{Cartpole Simulator}
\label{sec:cartpole}

\begin{figure}[t]
\centering
\includegraphics[width=0.47\textwidth]{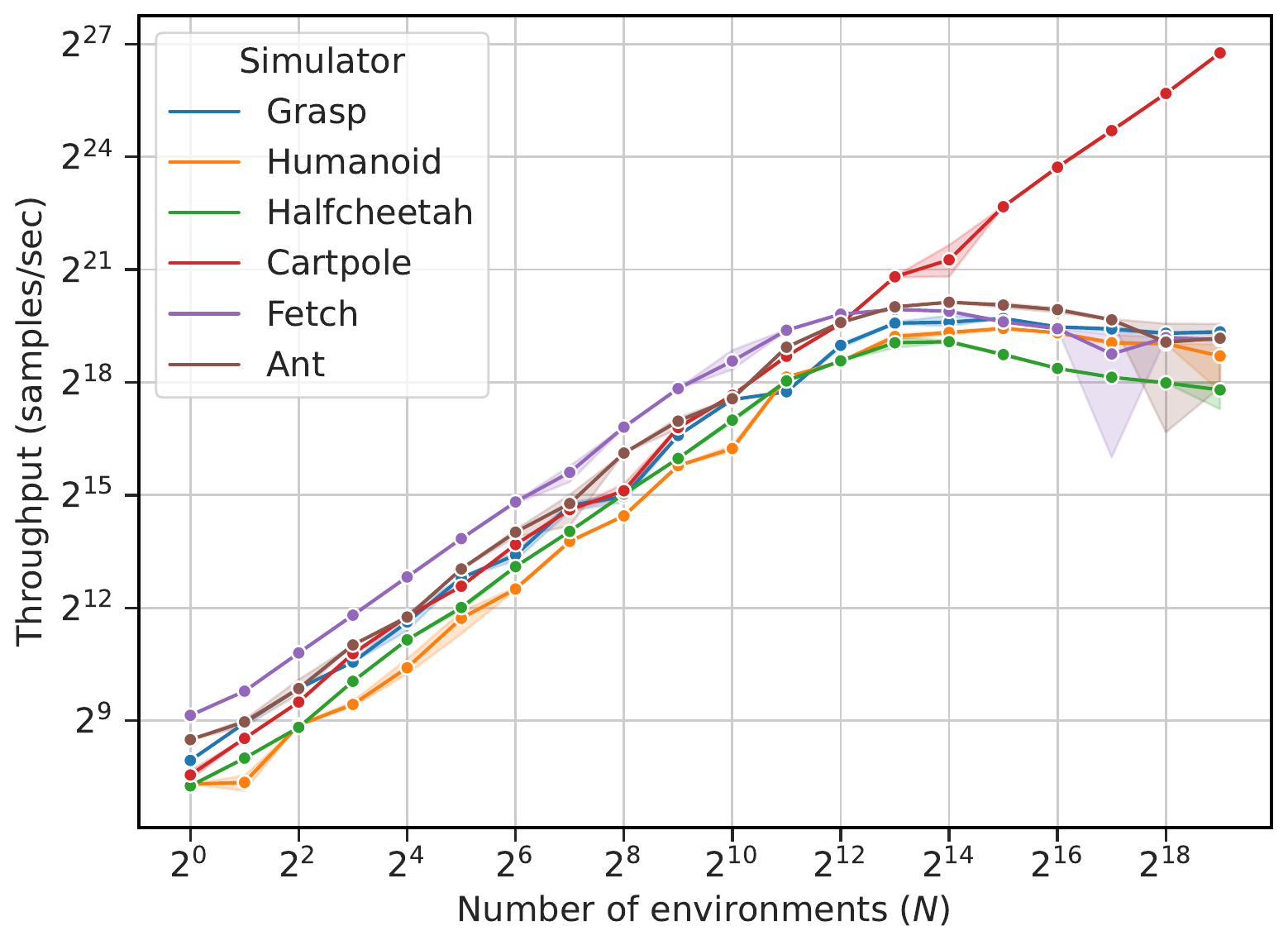}
\caption{\textbf{Comparing popular RL simulators:} 
simulation throughput of \CudaSimXLA GPU implementations of popular robotics physics simulators are shown.
The throughput of the simple cartpole simulator is similar to robotics
simulators, which allows us to focus our analysis on
cartpole to reduce implementation complexity.}
\label{fig:compare-simulators}
\end{figure}

\skeleton{
    \BulletPoint{Cartpole is a commonly used and computationally simple physics simulation consisting of 4 states and element-wise operations (\Cref{fig:data-collection}).}
    \BulletPoint{We focus on the cartpole simulator since it is a best-case for the optimizations this paper explores.}
    \BulletPoint{It is a best-case for \emph{\GpuVec} since we can maximize throughput of parallel simulators before exceeding GPU memory capacity.}
    \BulletPoint{It is a best-case for \emph{inference/simulation kernel fusion} since element-wise operations benefit being memory-bound and having short duration (e.g., activation functions).}
}{
The cartpole \cite{cartpole} simulation models a cart moving left and right along a 2D plane to balance a pole, as illustrated in \Cref{fig:data-collection}.
The simulation state size is small consisting of $S = 4$ floating point numbers: the cart's velocity ($\vec{v}$), position ($x$), the pole's joint angle relative to the cart ($\theta$), and pole's angular velocity ($\omega$).
The simulation dynamics equations consist of simple row-wise transformations that produce a new set of 4 floats.

In this study, we focus on the cartpole simulator since it is simple to implement in DNN frameworks and directly in CUDA \cite{cuda}, and its performance is very similar to other common robotics physics simulators.
To test this, we directly compared the simulation throughput of \CudaSimXLA GPU implementations of cartpole against commonly used robotics physics simulators.
From \Cref{fig:compare-simulators}, we observe that the simulation throughput of cartpole is similar to common robotics simulators used in RL.
The main difference is that simulation throughput for cartpole scale up to $2^{19}$ environments, whereas Brax \cite{brax2021github} environments only scale up to $2^{12}$ environments, which is attributable to the additional compute complexity of Brax simulations which also include collision detection.
To understand the effect of simulator complexity on \gpuVec and kernel fusion speedups, we also vary the compute and memory bandwidth requirements of the cartpole simulation in \Cref{sec:simulator-complexity}.

}

\subsection{ML Frameworks}
\label{sec:ml-frameworks}

RL developers have a variety of ML frameworks to choose from when implementing RL training algorithms.
While most RL developers choose ML frameworks solely based on convenience, in this paper we instead delineate ML frameworks by subtle design choices that affect their performance.
As we will see, these performance differences become more pronounced when we investigate optimizing data collection.

\textbf{DNN frameworks} provide a pre-compiled library of GPU kernels for common deep learning operators (e.g., matrix multiplication, convolution, etc.), allowing developers to compose and build model architectures.
GPU kernels are either provided by the framework itself (e.g., element-wise operations), or accelerator-specific vendor libraries that have tuned specific operations for high-performance (e.g., matrix multiplication operators from the cuBLAS \cite{cublas} library when using NVIDIA GPUs).
The DNN framework we study in this paper is PyTorch, which is popular for its developer-friendly eager execution model.
Eager execution executes operations as GPU kernels as they are issued from Python \cite{python}, which conveniently allows developers to inspect intermediate DNN computations.
While convenient, a greater number of GPU kernel launches can suffer from kernel launch overhead and increased memory bandwidth requirements from reading/writing device memory between kernel launches.

\textbf{ML compilers} parse the entire DNN computational graph and perform global optimizations over that graph to increase execution efficiency. 
In this paper, we focus on the XLA compiler \cite{xla}, which targets an intermediate representation consisting of elementary DNN operations which are amenable to execution on both TPU and GPU accelerators.
In particular, we use the JAX \cite{jax2018github} Python frontend library to XLA.
XLA provides an API analogous to high-level linear algebra libraries (i.e., Python's NumPy \cite{numpy}), but additionally supports just-in-time compilation to accelerator kernels using a symbolic tracing of tensor shapes.
As a result, certain simple operations like element-wise operations can be fused into a single GPU kernel launch to increase performance compared to DNN frameworks.

\section{\GpuVecCamel}
\label{sec:evaluation}

\begin{figure}[t]
\centering
\includegraphics[width=0.47\textwidth]{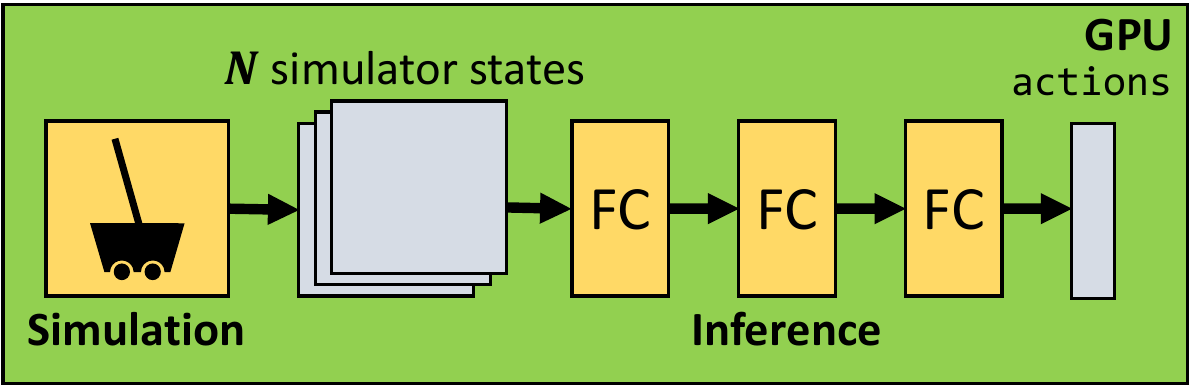}
\caption{
\textbf{Optimization 1. \gpuVec:} 
simulation is vectorized to operate on multiple parallel environments $N$ on the GPU. 
Multiple implementations can achieve this optimization, either through DNN operators (\CudaSimXLA, PyTorch), or by manually writing kernels to parallelize simulators across GPU threads (CUDA).
}
\label{fig:opt-1-vectorized-gpu-data-collection}
\end{figure}

\skeleton{
\BulletPoint{We compare Python based CPU simulators (like OpenAI gym, \Cref{fig:opt-1-vectorized-gpu-data-collection}) against vectorized GPU simulators}
}{
We study the performance limitations of \gpuVec (illustrated in \Cref{fig:opt-1-vectorized-gpu-data-collection}) through implementations spanning multiple ML frameworks: an ML compiler (\CudaSimXLA), and a DNN framework (PyTorch).
We demonstrate that both \CudaSimXLA and PyTorch have a \AsTimes{\CudaSimGpuOverCpuSpeedup} speedup when saturating parallel environments.
However, at many smaller configurations of parallel environments,
\CudaSimXLA outperforms PyTorch by \AsTimes{\CudaSimJaxDataCollectionSpeedupJaxOverTorch} by amortizing CPU overheads when combining multiple data collection steps in a single Python$\rightarrow$\CudaSimXLA API call.

}

\subsection{Hardware and Software Configuration}
\label{sec:hardware-software-config}

All analysis in this paper was performed on hardware consisting of 
an AMD EPYC 7371 CPU running at 3.1 GHz with 128 GB of RAM and an NVIDIA 2080Ti GPU.
For software configuration, 
we used Ubuntu 20.04, 
CUDA 11.2.0,
GCC 9.3.0,
PyTorch v1.8.1, 
JAX frontend v0.2.13 with \texttt{jaxlib} backend (XLA) v0.1.67, 
and Python 3.8.10.  

\subsection{Data Collection Implementations}
\label{sec:dl-data-collection-impls}

\skeleton{
    \BulletPoint{\textbf{OpenAI gym:} Uses a single CPU-core, is implemented in Python using numpy, and is very common in RL frameworks.  We still use PyTorch for inference.}
    \BulletPoint{\textbf{PyTorch:} we express multiple simulators instances efficiently on the GPU using vectorized DL operators that operate on a $N \times S$ state matrix.}
    \BulletPoint{\textbf{\CudaSimXLA:} same as PyTorch, except we reduce Python$\rightarrow$DL backend API calls that leads to more speed up.}
}{

To measure the benefit of \gpuVec, we created data collection implementations spanning multiple hardware types and ML frameworks. 
For the CPU implementations of simulation, we consider both the status-quo single-core CPU approach used by popular RL training frameworks (OpenAI gym), and a custom C++ implementation that utilizes multiple CPU cores.
For the GPU implementations of simulation, we consider commonly used DNN frameworks (PyTorch), and more performance-oriented ML compilers (\CudaSimXLA).

\textbf{OpenAI gym:} OpenAI gym provides a scalar (single instance) implementation of the cartpole simulator that stores the simulator state within a numpy array of $S = 4$ floating point state values.
Each step of the simulator is performed using an object-oriented API with a $step()$ state transition function, with multiple simulator instances implemented using multiple object instances.
Due to inefficient shared memory multi-threading in the Python high-level language, RL training frameworks opt to run multiple simulator instances on a single CPU core to maximize their performance.
MLP inference uses a single PyTorch model and combines multiple simulator outputs into a single inference minibatch that is copied to a GPU tensor input.
This common approach to data collection is illustrated in \Cref{fig:data-collection}.

\textbf{C++:} To explore the limits of multi-core CPU-based simulator implementations without being limited by the high-level language overheads of OpenAI gym, we implement the cartpole simulator using shared memory multi-threading.
While the simulation runs on the CPU, the inference component still uses PyTorch, with inference/simulation output being shared efficiently through a shared device memory allocation.

\textbf{PyTorch:} We converted the OpenAI gym cartpole simulator into a GPU implementation using PyTorch operators.
This is achieved by vectorizing the numpy implementation by storing multiple ($N$) simulator instances in an $N \times S$ state matrix, and replacing scalar operations with vectorized ones.
PyTorch is an eagerly executed DL framework; that is, Python$\rightarrow$DL backend API calls are executed on the GPU as the user invokes them from Python.

\textbf{\CudaSimXLA:} Similar to PyTorch, we created a vectorized GPU implementation using \CudaSimXLA operators by storing multiple ($N$) simulator instances in an $N \times S$ state matrix.
In contrast to PyTorch which eagerly executes Python$\rightarrow$DL backend calls on the GPU, JAX's numpy-like interface builds an entire symbolic computational graph composed of XLA operators.
This graph-based approach allows us to perform multiple data collection \emph{steps} in a single Python$\rightarrow$DL backend call which, as we will see, has performance implications.
}

\subsection{Data Collection Throughput}
\label{sec:gpu-data-collection}

\begin{figure}[t]
\centering
\includegraphics[width=0.47\textwidth]{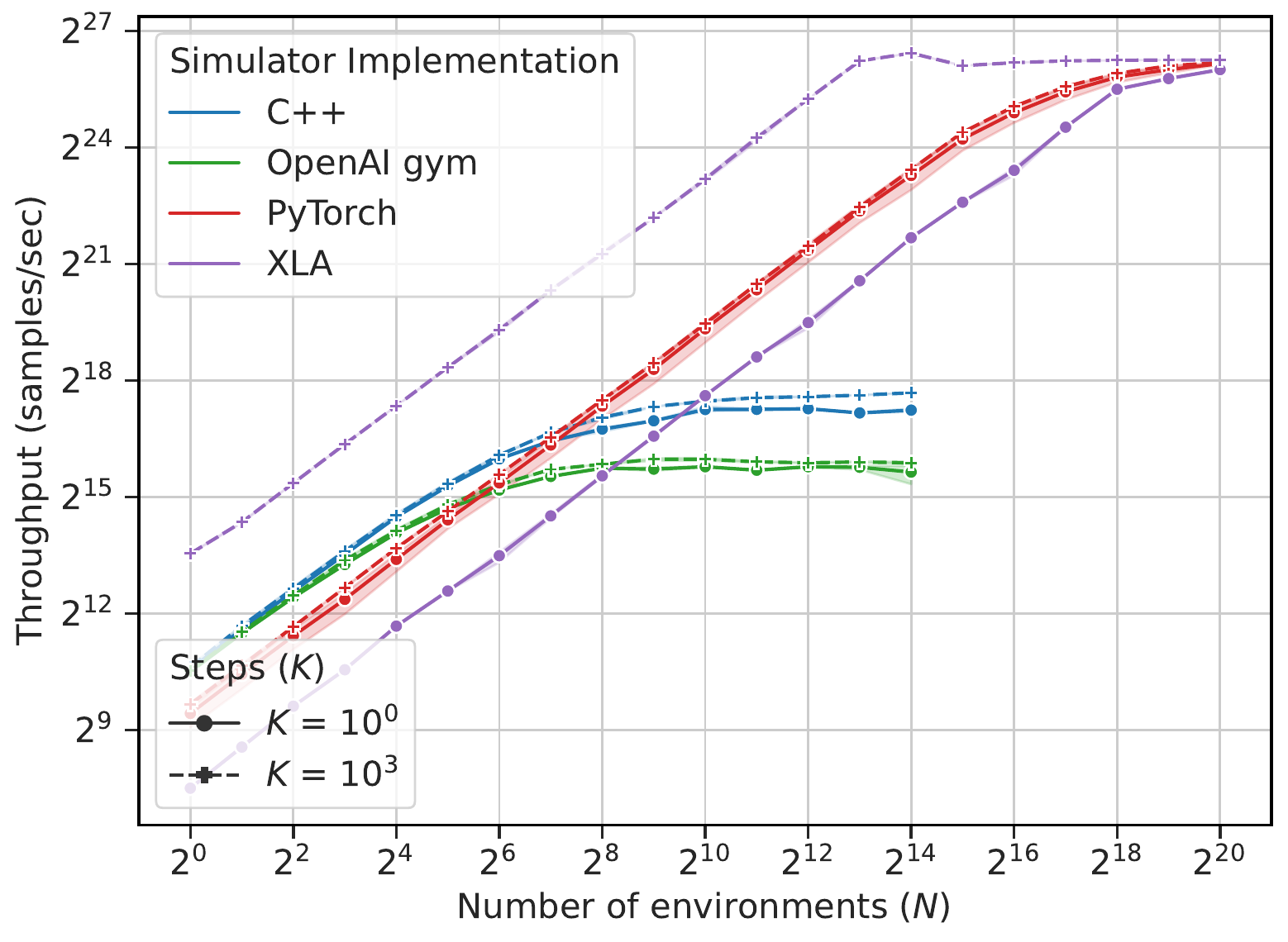}
\caption{\textbf{GPU data collection:} 
DL frameworks can express simulation as GPU computation,
exploiting GPU hardware parallelism and reducing CPU overheads from switching back and forth to GPU computation.
GPU implementations achieve 
much higher throughput than both CPU implementations (OpenAI gym, C++) of simulation.
}
\label{fig:skeleton-gpu-data-collection}
\end{figure}

\skeleton{
    \BulletPoint{All reach a data collection throughput plateau, with \CudaSimXLA and PyTorch both plateauing at $2^{16}$ samples/sec.}
    \BulletPoint{It remains unclear why \CudaSimXLA benefits from a greater number of data collection steps; to find out, in the next section we profile \CudaSimXLA and PyTorch implementations.}
}{
In \Cref{fig:skeleton-gpu-data-collection}, we run the data collection loop and measure the throughput in samples/second of data collected.
Each implementation has a maximum data collection throughput plateau:
C++ at $2^{18}$, 
OpenAI gym at $2^{16}$, 
PyTorch at $2^{26}$, 
\CudaSimXLA at $2^{26}$.
Moving simulation to the GPU using either PyTorch or \CudaSimXLA can provide up to a \AsTimes{\CudaSimGpuOverCpuSpeedup} speedup over the status quo approach of OpenAI gym's CPU simulation.
Using multiple CPU cores with C++ only accounts for at most a 
\AsTimes{\CudaSimDataCollectionMultiCoreCpuSpeedup} 
speedup over the single-core OpenAI implementation.
}

\begin{finding}{find:gpu-data-collection-speedup}
Both DL framework (PyTorch) and ML compiler (\CudaSimXLA) \gpuVec approaches can provide up to a $\CudaSimGpuOverCpuSpeedup\times$ speedup in data collection over OpenAI gym.
\end{finding}

\skeleton{
    \BulletPoint{\Cref{fig:skeleton-gpu-data-collection} shows data collection throughput for OpenAI, PyTorch, and \CudaSimXLA implementations.}
    \BulletPoint{As the number of data collections steps increases, \CudaSimXLA amortizes overheads associated with calling from Python$\rightarrow$DL backend.}
    \BulletPoint{PyTorch cannot benefit from multiple steps, since each step results in separate Python$\rightarrow$DL backend calls.}
}{
Since \CudaSimXLA can express loop control-flow structures, we can define the entire data collection loop as a computational graph, which is invoked with a single Python$\rightarrow$\CudaSimXLA API call. 
Since PyTorch is eagerly executed, control-flow constructs are expressed in Python resulting in multiple calls from Python into PyTorch for each step of data collection.
As a result, the data collection throughput for PyTorch does not change as we increase the steps.
On the other hand, \CudaSimXLA benefits from increased throughput.
}

\skeleton{
    \BulletPoint{Increasing steps $K$ increases \CudaSimXLA performance.}
    \BulletPoint{To explain this performance increase, in the next section we profile \CudaSimXLA.}
}{
Increasing the number of consecutive data collection steps executed in a single Python$\rightarrow$\CudaSimXLA API call from $10^0\rightarrow10^3$ results in a 
\AsTimes{\CudaSimJaxDataCollectionSpeedupSteps} speedup on average for most environment sizes ($\leq 2^{13}$), which is a \AsTimes{\CudaSimJaxDataCollectionSpeedupJaxOverTorch} speedup over PyTorch.
An important question this raised was whether \CudaSimXLA could be performing kernel fusion across multiple steps, given that it has complete computational graph knowledge.
To investigate this, we performed a profiling deep dive of the \CudaSimXLA implementation.
}

\subsection{Time Breakdown of \CudaSimXLA GPU Data Collection}
\label{sec:time-breakdown}

\skeleton{
    \BulletPoint{We used RL-Scope to get a CPU/GPU time breakdown across inference and simulation.}
}{
To understand the underlying bottlenecks that exist in the \CudaSimXLA GPU implementation of data collection, it is helpful to get a time breakdown of where total execution time is spent across the stack.
To obtain this time breakdown, we used RL-Scope \cite{gleeson2021rlscope}.  RL-Scope provides a full-stack breakdown of time across the CPU, GPU, and within different parts of RL computation, such as inference and simulation.
}

\begin{figure}[t]
\centering
\includegraphics[width=0.35\textwidth]{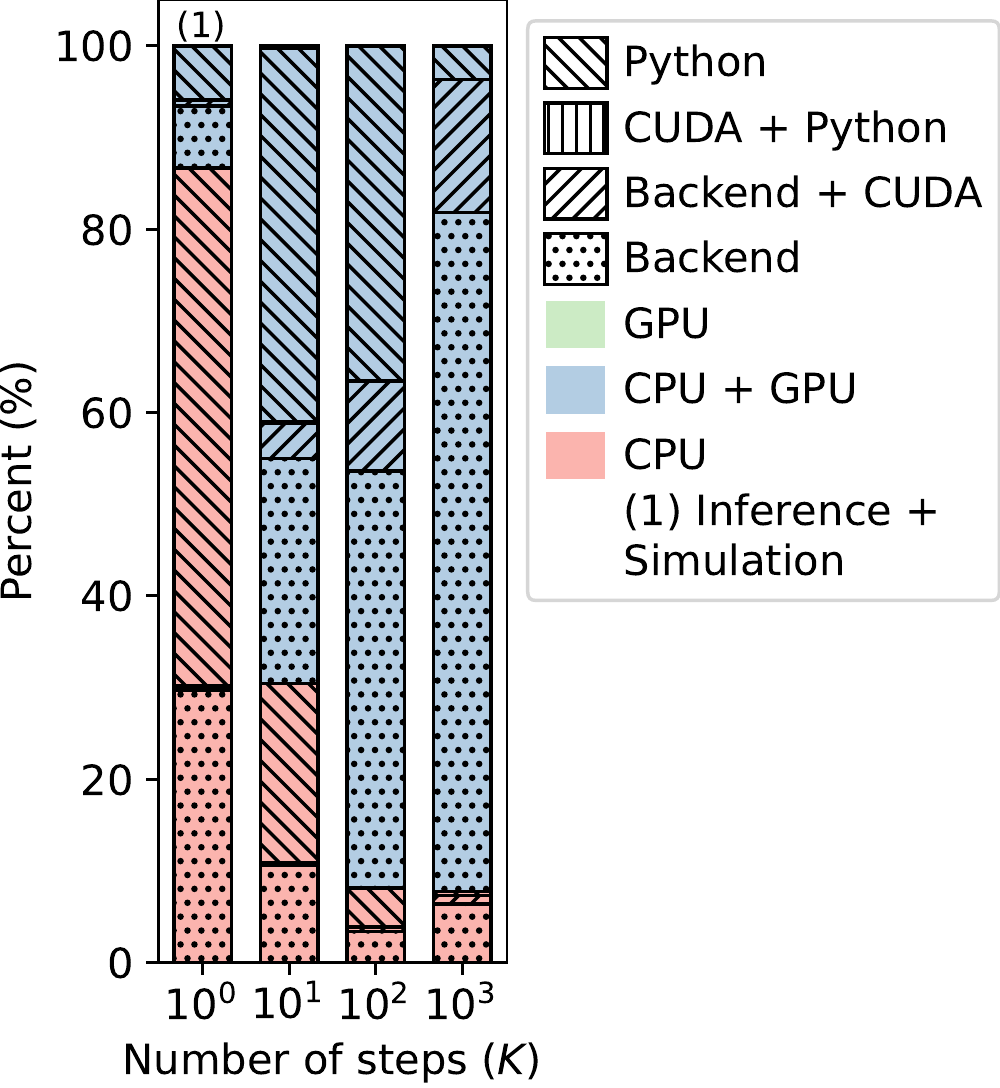}
\caption{\textbf{\CudaSimXLA GPU time breakdown:} 
as we increase the number of data collection steps completed in a single Python$\rightarrow$XLA API call, CPU overheads from Python and Backend C++ code are amortized, and GPU time saturates the computation.
Number of environments $N = 2^{16}$.}
\label{fig:skeleton-time-breakdown-jax}
\end{figure}

\begin{finding}{find:time-breakdown-gpu-data-collection-no-fusion}
In contrast to DNN frameworks (e.g., PyTorch),
ML compilers that can express control-flow constructs (e.g., \CudaSimXLA) can reduce CPU-based overhead by reducing Python$\rightarrow$Backend API calls. 
However, more complete computational graph information is not being used to fuse simulation and inference GPU kernels.
\end{finding}

\begin{figure}[t]
\centering
\includegraphics[width=0.33\textwidth]{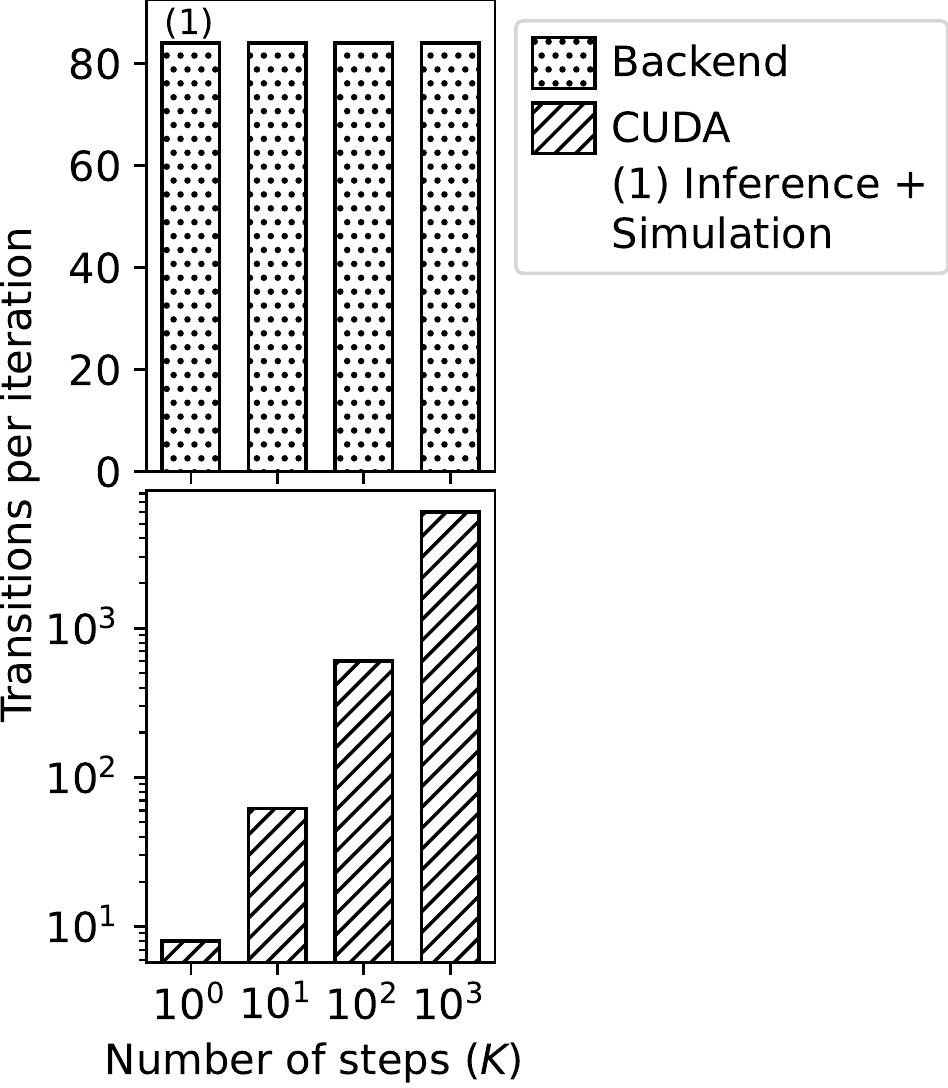}
\caption{\textbf{\CudaSimXLA API transition counts:} 
\emph{CUDA} shows the number of GPU kernel launches, and \emph{Backend} shows the number of Python$\rightarrow$DL backend (i.e., \CudaSimXLA) API calls.
Python$\rightarrow$XLA API calls (\emph{Backend}) remain constant even as we increase data collection steps, explaining amortized CPU overheads.
However, GPU kernel launches (\emph{CUDA}) increase linearly, indicating kernel fusion does not occur across data collection steps.
}
\label{fig:skeleton-transition-count}
\end{figure}

\skeleton{
    \BulletPoint{Increasing data collection steps with \CudaSimXLA amortizes serialization/deserialization overheads for Python$\rightarrow$DL backend calls (\Cref{fig:skeleton-time-breakdown-jax}).}
    \BulletPoint{However, \CudaSimXLA does not make any attempt to fuse simluation/inference GPU kernels calls, since they scale linearly with increasing data collection steps (\Cref{fig:skeleton-transition-count}).}
    \BulletPoint{In the next section, we will investigate the potential speedup from fusing simulation/inference GPU kernels.}
}{
In \Cref{fig:skeleton-time-breakdown-jax}, we observe that at small step sizes like $10^{0}$, CPU-bound operations 
make up $87\%$ of data collection time split across backend C++ calls ($30\%$) and Python time ($57\%$).
As we increase the number of steps executed in a single \CudaSimXLA API call, CPU overheads are amortized, which contributes to the overall speedup achieved by the \CudaSimXLA implementation at higher step sizes.

To determine whether these performance benefits could be due to kernel fusion by \CudaSimXLA,
we measured the number of GPU kernel launches during an iteration of data collection.
In \Cref{fig:skeleton-transition-count}, 
\emph{CUDA} shows the number of GPU kernel launches, and \emph{Backend} shows the number of Python$\rightarrow$DL backend (i.e., \CudaSimXLA) API calls.
In \Cref{fig:skeleton-transition-count}, as expected, as we increase the number of steps handled in a single \CudaSimXLA API call, the number of Python$\rightarrow$Backend calls remains constant, explaining why CPU overhead was amortized.
If we look at the number of GPU kernel launches (\emph{CUDA}), 
 they scale linearly with the step size.  
This tells us that \CudaSimXLA is not performing kernel fusion at higher step sizes.
Further inspection of kernel launches reveals additional calls to simulator kernels, and also cuBLAS matrix multiplication kernels from inference which are inherently infusible since they are closed source.

Our profiling demonstrates that kernel fusion remains an unexplored optimization for the RL training loop.
As an initial step in this direction, we next investigate fusing multiple simulation steps 
to understand the potential performance benefit of fusion.

}

\section{\SimKernelFusionCamel}
\label{sec:kernel-fusion}

\begin{figure}[t]
\centering
\includegraphics[width=0.30\textwidth]{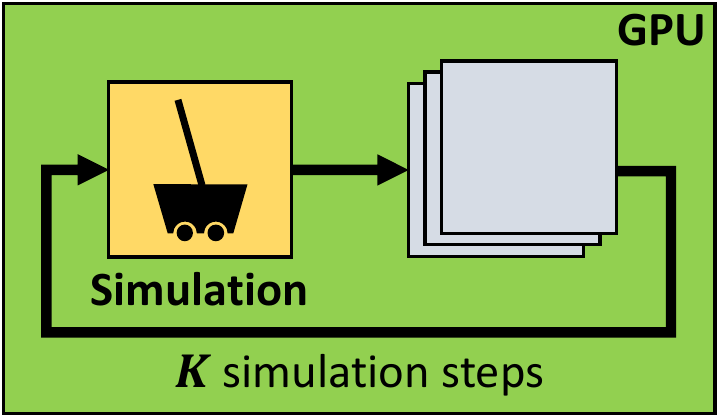}
\caption{\textbf{Optimization 2. \SimKernelFusion:} 
simulation is a short duration kernel that performs row-wise operations, so fusing it into one kernel launch in CUDA will benefit from (1) reduced launch overhead from fewer kernel launches, and (2) increased cache efficiency of simulator state.
$K$ simulator steps are executed in a single fused GPU kernel launch.
}
\label{fig:opt-2-simulation-inference-fusion}
\end{figure}

\skeleton{
\BulletPoint{In this section, we explore the potential speedup in data collection that is possible from fusing the simulation kernel with the first layer of inference, as illustrated in \Cref{fig:opt-2-simulation-inference-fusion}.}
}{
GPU kernel fusion can benefit from (1) reduced kernel launch overhead from fewer kernel launches, and (2) increased cache efficiency by avoiding device memory transfers. 
Based on our profiling results (\Cref{sec:time-breakdown}), more performant ML compiler implementations cannot fuse GPU kernel launches across multiple data collection steps.
To study the effects of kernel fusion, we implement a purely CUDA implementation of simulation where we can precisely control the number of GPU kernel launches during simulation (illustrated in \Cref{fig:opt-2-simulation-inference-fusion}).
We demonstrate that kernel fusion is an orthogonal optimization that can be combined with \gpuVec leading to multiplicative benefits.
}

\subsection{\SimKernelFusionCamel Throughput}
\label{sec:fused-simulation}

\skeleton{
    \BulletPoint{We estimate an upper bound on inference/simulation kernel fusion by fusing multiple simulation steps only.}
    \BulletPoint{As before we consider PyTorch, \CudaSimXLA, and OpenAI gym implementations of multi-step simulation.  \CudaSimXLA/PyTorch execute multiple kernels for multiple data collections steps.  In addition, we consider:}
    \BulletPoint{\textbf{CUDA:} this maximizes GPU throughput by keeping simulator state entirely in registers.  A single kernel launch executes all data collection steps.}
    \BulletPoint{\textbf{C++:} this maximizes CPU throughput using efficient shared memory multi-threading, in contrast to OpenAI gym (Python).}
}{

Since the simulation kernel for cartpole is short and consists of simple row-wise operations, fusing simulation steps will benefit both from reduced launch overhead and increased efficiency of keeping intermediate simulator state in cache.
To explore kernel fusion, we added a CUDA implementation, in addition to the GPU-based PyTorch and \CudaSimXLA
implementations of simulation 
(\Cref{sec:dl-data-collection-impls}):

\textbf{CUDA:} To explore the limits of GPU-based kernel fusion and control kernel fusion decisions, we manually implemented a GPU kernel for simulating multiple cartpole instances in parallel.  
Multiple simulation steps are executed within a single kernel launch, minimizing kernel launch overhead and maximizing data reuse.

}

\begin{figure}[t]
\centering
\includegraphics[width=0.47\textwidth]{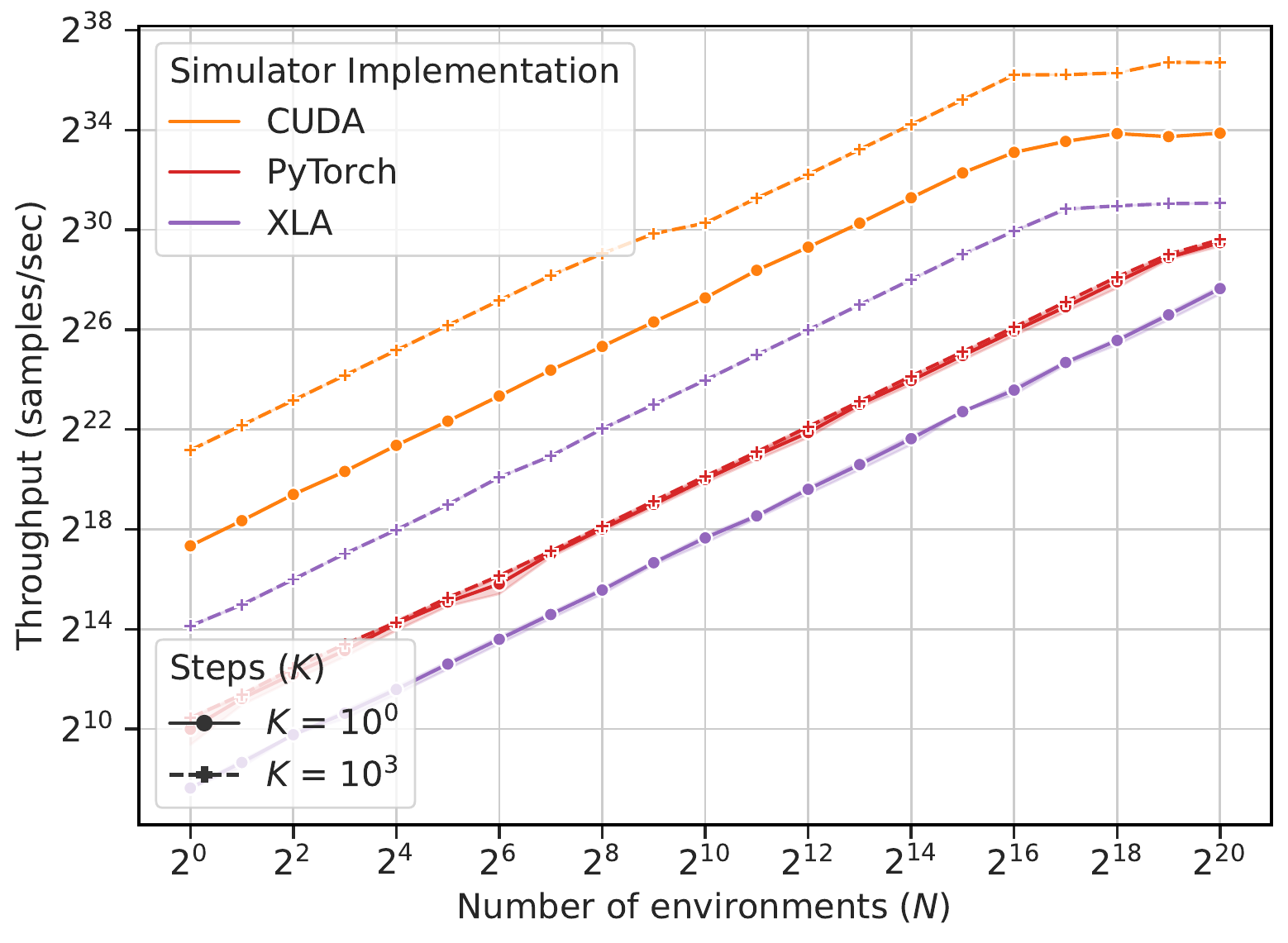}
\caption{\textbf{\SimKernelFusion:} simulation throughput benefits from increased fusion at a higher number of fused simulation steps $K$.  \CudaSimXLA does not fuse GPU kernels but still benefits from amortizing CPU overheads.  CUDA benefits from fusing kernel launches (i.e., reduced launch overheads, cached simulator state).}
\label{fig:skeleton-fused-simulation-steps}
\end{figure}

\skeleton{
    \BulletPoint{In \Cref{fig:skeleton-fused-simulation-steps}, we measure the simulation throughput of different implementations of multi-step simulation.}
    \BulletPoint{As data collection steps increase, the CUDA implementation benefits more from kernel fusion, attributable to reduced launch overheads and caching simulator state in registers.}
    \BulletPoint{Next, we investigate whether the full data collection loop can benefit from similar speedups observed in multi-step simulation.}
}{
In \Cref{fig:skeleton-fused-simulation-steps}, we show the simulation throughput of different implementations of multi-step simulation.
Since PyTorch has an eager API, the number of GPU kernel launches scales linearly with the number of steps and cannot benefit from kernel fusion, causing the $K = 10^0$ and $K = 10^3$ lines of PyTorch to overlap.
Since \CudaSimXLA is provided with the full computational graph for iteratively performing simulation, multiple simulation steps can be condensed into a single Python$\rightarrow$\CudaSimXLA call, amortizing CPU overheads, leading to a \AsTimes{\CudaSimSimulationJaxAmortizedCpuOverheadSpeedup} speedup for \CudaSimXLA going from $K = 10^0 \rightarrow 10^3$.

Since CUDA is handwritten to execute multiple simulator steps in a single kernel launch, it can benefit both from reduced kernel launch overhead and better caching of simulator state in registers.
The \AsTimes{\CudaSimSimulationKernelFusionAndNoCpuOverheadSpeedup} speedup for CUDA$\rightarrow$\CudaSimXLA for $K = 10^3$ benefits both from kernel fusion and reduced framework API overhead.
To isolate the benefit from kernel fusion alone, we can consider the speedup for CUDA going from $K = 10^0 \rightarrow 10^3$ steps; we see that kernel fusion accounts for a \AsTimes{\CudaSimSimulationKernelFusionSpeedup} speedup.
}

\begin{finding}{find:fused-simulation}
For repeated simulation steps, kernel fusion can provide \AsTimes{\CudaSimSimulationKernelFusionSpeedup} speedup over an unfused implementation of the cartpole simulator.
\end{finding}

\subsection{Simulator Complexity}
\label{sec:simulator-complexity}

\begin{figure*}[t]
\centering
\includegraphics[width=0.98\textwidth]{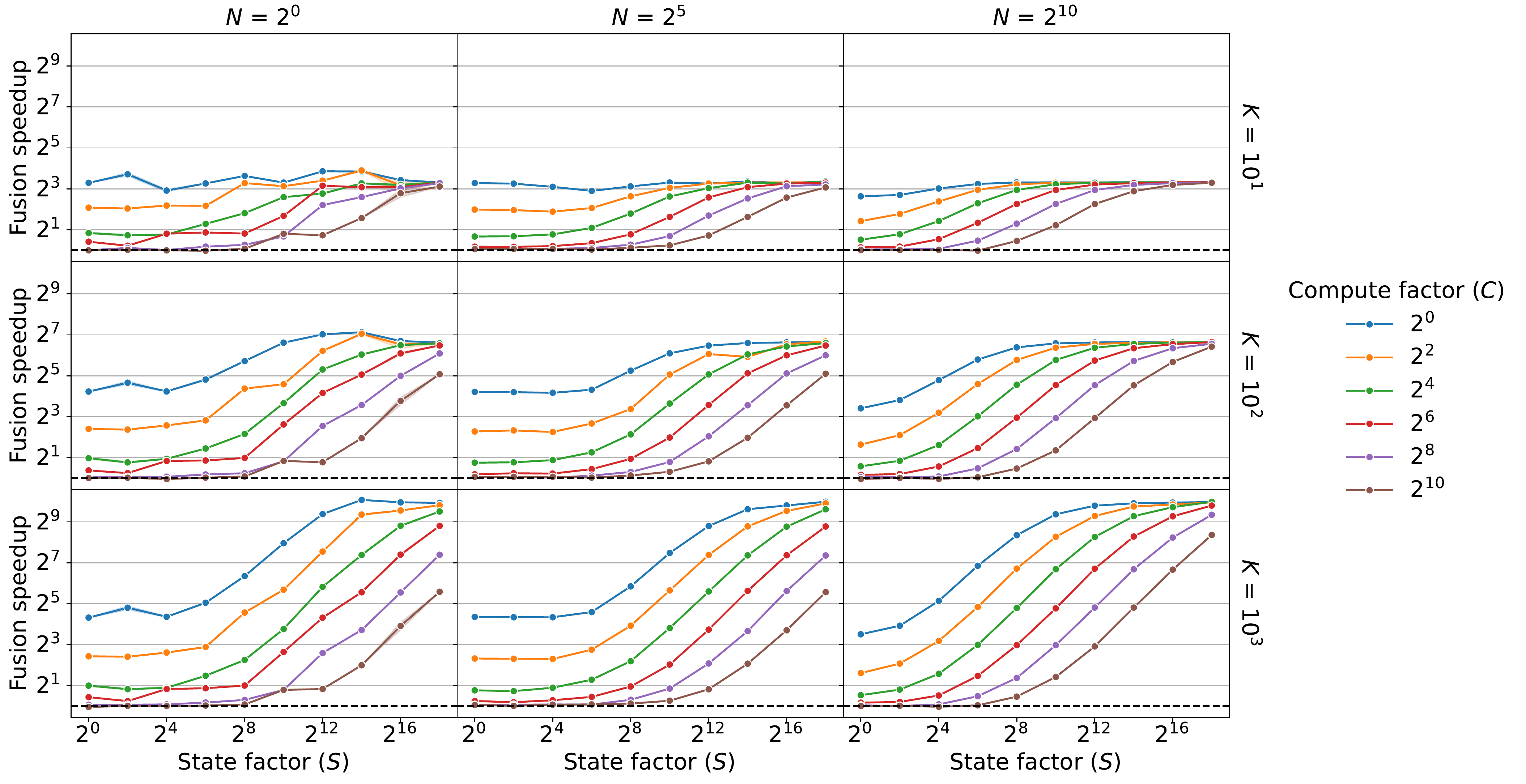}
\caption{\textbf{Simulator complexity:} the compute factor $C$ increases cartpole compute requirements by $C\times$ and state factor $S$ increases memory bandwidth requirements by $S\times$.  The fusion speedup is measured by normalizing with respect to a fusionless run (i.e., steps $K = 10^{0}$).
Parallel environments $N$ increase from left to right subplots, and number of fused steps $K$ from top to bottom subplots.
}
\label{fig:simulator-complexity}
\end{figure*}

An important consideration is how the speedup from fusion varies as we increase the complexity of the simulator.
In particular, the simulator could be more compute-bound, more memory bandwidth bound, or both.
To answer this question, we took the cartpole simulator and added configurable parameters to increase the compute and memory requirements of the simulation.
The compute factor $C$ increases the compute requirements of cartpole by $C \times$ by increasing the number of simulation iterations.
The state factor $S$ increases the cartpole state size by a factor of $S \times$, resulting in $S \times$ as many global memory loads and stores.
To measure the speedup due to fusion, we normalize the simulation throughput with respect to a fusionless run (i.e., steps $K = 10^{0}$).
\Cref{fig:simulator-complexity} shows the speedup from fusion along all configurable workload dimensions 
(parallel environments $N$, fused steps $K$, state factor $S$, compute factor $C$).

\textbf{Number of parallel environments ($N$)}:
As we increase $N$ (left to right across subplots), the speedup from fusion remains the same.
For example, the last row of plots (steps $K = 10^{3}$) all plateau to a fusion speedup of $2^{10}$.
Hence, the benefit from kernel fusion is independent of how many parallel environments are used; 
kernel fusion is an orthogonal and combinable optimization with \gpuVec.

\textbf{State size factor ($S\times$):}
As we increase $S$ (left to right along the x-axis of each subplot), the simulation becomes increasingly memory bandwidth bound due to increased global memory loads/stores.
Increasingly memory bandwidth bound kernels benefit more from fusion since fusion reduces global memory transfers.
For example, for the last row of plots (steps $K = 10^{3}$), the speedup varies from $2^{4}$ to $2^{10}$ as $S$ increases.

\textbf{Number of fused steps ($K$):}
As we increase $K$ (top to bottom across subplots) the speedup increases more significantly in memory bandwidth bound workloads.
For example, 
less fusion (steps $K = 10^{1}$) achieves at most $2^{3}$ speedup, 
whereas more fusion (steps $K = 10^{3}$) achieves at most $2^{10}$ speedup at large state factors ($S \ge 2^{11}$).

\textbf{Compute factor ($C$):}
As we increase $C$ (the lines of each subplot), the simulation becomes increasingly compute-bound due to additional simulation iterations.
The more compute-bound the workload is, the less speedup that comes from fusion since the workload is not benefitting from greater data reuse by reducing global memory transfers.
In the worst case where the simulation is entirely compute bound ($C = 2^{10}$) and not memory bandwidth bound ($S = 2^{0}$), fusion performs just as poorly as without fusion.

\begin{finding}{find:simulator-complexity}
Kernel fusion benefits memory bandwidth bound simulation kernels with low to moderate compute complexity by reducing global memory transfers.
The speedup from kernel fusion is independent of how many parallel simulators are used 
and should be considered an orthogonal and hence combinable optimization.
\end{finding}

\section{Related Work}
\label{sec:related-work}

Related work falls into two categories, which we summarize here and discuss in detail below.
Prior work has demonstrated training time speedups from moving simulation to the GPU by performing \emph{\gpuVec} either by expressing simulation as DNN operators \cite{brax2021github} or with manually written CUDA code \cite{cule,large-batch-sim}. 
In contrast to those works, we perform an apples-to-apples comparison of multiple vectorization implementations to understand inherent performance limitations associated with hardware and ML framework choice.
We demonstrate that ML compiler frameworks can outperform DNN framework implementations by amortizing CPU overheads, but that current ML compiler frameworks cannot perform kernel fusion that would benefit RL training workloads. 
\emph{Kernel fusion} is a common optimization pass for ML compilers but is typically limited to element-wise operations \cite{chen2018tvm}, with some compilers opting to offload high-performance operations to infusible cuBLAS libraries \cite{xla}.
In this paper, we do not limit ourselves to the current capabilities of ML compilers, and instead measure the potential benefits achievable if kernel fusion were implemented today in these ML compilers by exploring manual fusion of simulation kernels implemented in CUDA.

\textbf{\gpuVec:}
Brax \cite{brax2021github} implements a general physics engine in JAX that can support many environments consisting of connected rigid bodies, allowing them to support common robotics physics simulators.
The physics engine implementation parallelizes the computation across both environments and joint angles.  
Brax leverages accelerator (i.e., GPU, TPU) parallelism during simulation in RL training, allowing them to achieve significant training time speedups ($266\times$ on a V100 GPU) 
over simple single CPU core simulators \cite{todorov2012mujoco}.
In contrast to Brax, our evaluation of the performance benefit of \gpuVec is more thorough for four reasons.
First, we consider a multi-core C++ implementation for the CPU instead of just a naive single-core CPU implementation.
Second, we demonstrate a \AsTimes{\CudaSimJaxDataCollectionSpeedupSteps} speedup in \CudaSimXLA by combining multiple data collection steps in a single Python$\rightarrow$\CudaSimXLA API call to amortize CPU overheads.
Third, besides an ML compiler approach (\CudaSimXLA), we also compare against common DL framework implementations (PyTorch) and demonstrate how the superior \AsTimes{\CudaSimJaxDataCollectionSpeedupJaxOverTorch} performance of \CudaSimXLA over PyTorch on the GPU is achieved through multiple data collection steps.
Finally, through profiling, we show that \CudaSimXLA cannot perform kernel fusion in data collection, which motivates us to investigate kernel fusion as a novel and orthogonal optimization in RL training.

CuLE \cite{cule} explores porting the C++ Atari emulator to CUDA to run on GPUs, with each GPU thread executing an Atari hardware emulator instance.
Emulating the Atari CPU architecture on the GPU leads to degradation in performance due to branch divergence.
As a result, the observed speedup over CPU is at most $2.54 \times$ when training Atari Assault on a Titan V GPU.
In contrast to this paper, CuLE implements the Atari simulation only in CUDA; 
it is likely not possible to express non-data-parallel operations like hardware instruction emulation using DNN framework or ML compiler supported operators.
We instead focus on physics simulators that are more amenable to the GPU architecture, allowing us to 
investigate the performance limitations of both DNN frameworks and ML compiler implementations, 
and explore kernel fusion optimizations that are not explored in CuLE.

\textbf{Kernel fusion:}  TVM \cite{chen2018tvm} is an ML compiler that enables developers to write DNN operators in a high-level domain-specific language (DSL), with the final output being lowered to various different hardware backends (e.g., GPU, TPU, FPGA).
TVM implements several optimization passes over the computational graph, including operator fusion.
TVM operator fusion is limited to fusing with element-wise operations (called ``injective'' operators) as they are the simplest to fuse.
In particular, injective operators fuse with: other injective operators (e.g., $BatchNorm \circ relu$), the input of a reduction (e.g., $scale \circ sum$), or the output of a complex-out-fusible (e.g., $conv2d \circ relu$).
We chose not to use TVM to implement fusion since our simulator is row-wise not element-wise, and we wanted to tightly control how the simulator state is managed during fusion 
and explore performance implications.

XLA \cite{xla} is a compiler for linear algebra programs (e.g., DNNs) with operators that can be lowered to backend targets (i.e., TPU, GPU, CPU).
Frontend APIs like JAX \cite{jax2018github} builds on top of XLA by offering JIT compilation of operator graphs into optimized backend code.
The operator fusion compiler pass heuristically tries to reduce memory bandwidth \cite{xla-gpu-kernel-fusion}.
Producer-consumer fusion allow fusion of element-wise operations (e.g. fused $add \circ relu$).
Sibling fusion fuses operators that share a common input to reduce global memory reads, interleaving the operator outputs as a tuple
(e.g., fusing $sum(x)$ and $sum(x^2)$ for computing $BatchNorm$).
In this paper, we found that kernel fusion across the inference and simulation boundary is not possible since XLA still relies on closed-source cuBLAS matrix multiplication kernel implementations.
This motivated us to explore the potential benefits of performing kernel fusion in the simulation phase of training by manually implementing simulation CUDA kernels, and how these speedups vary with respect to simulator complexity.
Our exploration of the circumstances under which simulators benefit from kernel fusion are important for incorporating compiler-based kernel fusion optimizations that target RL workloads.

\section{Conclusions}
\label{sec:conclusions}

We demonstrated that the speedups from \gpuVec are substantial, with up to \AsTimes{\CudaSimGpuOverCpuSpeedup} speedup for a simple simulator.
We showed that ML compilers (XLA) outperform DNN frameworks (PyTorch) by \AsTimes{\CudaSimJaxDataCollectionSpeedupJaxOverTorch} by amortizing CPU overheads across multiple backend API calls.
For \simKernelFusion, we showed that the speedups from kernel fusion for a simple simulator are \AsTimes{\CudaSimSimulationKernelFusionSpeedup} and increase by up to \AsTimes{\CudaSimSimulatorComplexityMaxSpeedup} as simulator complexity increases in terms of memory bandwidth requirements.
The speedups from kernel fusion are orthogonal and combinable with \gpuVec, leading to a multiplicative speedup.  
We hope our study spurs greater interest in specialized optimizations targeting emerging RL workloads.
\section{Future Work}
\label{sec:future-work}

Our initial analysis demonstrated that large speedups are possible with both \gpuVec and \simKernelFusion, and that these speedups are combinable.
In future work, we will study additional simulators such as robotics physics simulations, and perform fusion across simulation and inference GPU kernels to speedup the data collection process in today's RL workloads.

\textbf{Simulation/inference fusion:}
We limited our implementation of kernel fusion in CUDA to multiple simulation steps, but did not include inference.  
The benefit of this approach is that it reduced engineering complexity and allowed us to compare the performance and inherent limitations of several approaches to optimizing multi-step simulation 
(\CudaSimXLA, PyTorch, CUDA).
However, to accelerate RL training, in the future we must apply fusion to the full data collection simulation/inference loop.
Given that we cannot manually fuse closed-source GPU kernels, we will need to obtain high-performance open-source GPU kernels with performance comparable to cuBLAS.
Based on our analysis of large fusion benefits in memory bandwidth bound simulators, we suspect that kernel fusion will still benefit DNN inference computations that have memory bandwidth bound behaviour. 
Memory bandwidth bound kernels are known to occur in matrix multiplication for irregularly shaped tall/skinny matrices \cite{cho2021accelerating} that are typical for RL inference due to the simulator state matrix.

\textbf{Exploring additional simulators:}
Our analysis focused on the simple cartpole simulator, which allowed us to explore the influence of ML framework and hardware on kernel fusion across different implementations.
Further, our study of simulator complexity demonstrated that increasingly memory bandwidth bound simulators benefit the most from kernel fusion.
Hence, our immediate next step will be to explore additional simulators of varying memory/compute complexity to investigate how many existing simulators can benefit from kernel fusion.
Our first step in this direction will be to re-implement popular robotics simulators from Brax \cite{brax2021github} in CUDA and see how manual kernel fusion compares to the kernels produced by XLA.
We would also like to explore photorealistic simulators built on industrial video game engines used in autonomous driving \cite{nvidia-drive-sim} and factory robotics scenarios \cite{nvidia-isaac-sim} since these simulators are large contributors to total training time.

\bibliographystyle{ACM-Reference-Format}
\bibliography{rl_optimization_paper}

\end{document}